\pdfoutput=1

\documentclass[11pt]{article}

\usepackage{authblk}
\usepackage[]{ACL2023}

\usepackage{times}
\usepackage{latexsym}
\usepackage{amsmath}
\usepackage{graphicx}
\usepackage{tabularx}
\usepackage[normalem]{ulem}
\useunder{\uline}{\ul}{}

\usepackage[T1]{fontenc}

\usepackage[utf8]{inputenc}

\usepackage{microtype}

\usepackage{inconsolata}

\usepackage{makecell}

\title{UMLS-KGI-BERT: Data-Centric Knowledge Integration in Transformers for Biomedical Entity Recognition}

\author[1,2]{Aidan Mannion}
\author[3]{Thierry Chevalier}
\author[1]{Didier Schwab}
\author[1]{Lorraine Goeuriot}

\affil[1]{Univ. Grenoble Alpes, CNRS, LIG, Grenoble, France}
\affil[2]{EPOS SAS, Issy-les-Moulineaux, France}
\affil[3]{UFR de Médecine Univ. Grenoble Alpes, La Tronche, France}

\begin{document}
\maketitle
\begin{abstract}
Pre-trained transformer language models (LMs) have in recent years become the dominant paradigm in applied NLP.
These models have achieved state-of-the-art performance on tasks such as information extraction, question answering, sentiment analysis, document classification and many others.
In the biomedical domain, significant progress has been made in adapting this paradigm to NLP tasks that require the integration of domain-specific knowledge as well as statistical modelling of language.
In particular, research in this area has focused on the question of how best to construct LMs that take into account not only the patterns of token distribution in medical text, but also the wealth of structured information contained in terminology resources such as the UMLS.
This work contributes a data-centric paradigm for enriching the language representations of biomedical transformer-encoder LMs by extracting text sequences from the UMLS.
This allows for graph-based learning objectives to be combined with masked-language pre-training.
Preliminary results from experiments in the extension of pre-trained LMs as well as training from scratch show that this framework improves downstream performance on multiple biomedical and clinical Named Entity Recognition (NER) tasks.
All pre-trained models, data processing pipelines and evaluation scripts will be made publicly available.
\end{abstract}

\section{Introduction}
In recent times, transformer language models \cite{vaswani_attention_2017} have become the most popular and effective sequence modelling framework in almost all areas of applied Natural Language Processing.
Unsupervised pre-training on large quantities of text allows transformers to capture rich semantic and syntactic patterns that can be transferred to many specialised language processing objectives.
As such, transformer models that use the transfer learning paradigm whereby the model is trained in an unsupervised manner on a large text corpus and then fine-tuned on a downstream supervised-learning task have achieved state-of-the-art results across a wide range of general and domain-specific applications.

The proliferation of textual data in the biomedical domain (Electronic Health Records (EHRs), clinical documents, pharmaceutical specifications, etc) has precipitated the broad adoption of deep learning \& NLP techniques for information extraction and processing \citep{li_neural_2021,tiwari_assessment_2020,dubois_effectiveness_2017}.
Moreover, it has been shown that language models are capable of encoding clinical knowledge to a certain extent \cite{singhal_large_2022}.
Biomedical and clinical NLP, however, is widely recognised to present particular challenges that do not apply to the same extent in other domains, in particular the need to incorporate structured domain knowledge into text encodings \cite{chang_benchmark_2020}.
In order for neural language modelling to be reliable in a discipline as highly specialised as medicine, there is a more acute need for models to learn directly from domain-specific terminologies, as opposed to relying solely on corpus-based learning.
Thus, a significant amount of research effort in the medical NLP community has been directed towards the question of how best to inject information from knowledge graphs (KGs) into LMs \cite{he_kgmttbert_2022,naseem_incorporating_2022,li_behrt_2020}.
However, a generalisable, widely-accepted approach to this technique that can be easily transferred across different problem settings, models and training corpora has yet to emerge.
In addition, research into knowledge graph integration in NLP in the biomedical domain has tended to focus on English-language corpora; the utility and transferability of these techniques for other languages, for which less textual resources are available, as well as for multilingual models, remains therefore an under-explored area. 

This paper aims to contribute to the resolution of these issues by proposing a general framework for training BERT encoders \cite{devlin_bert_2019} using the UMLS (Unified Medical Language System, \citet{umls_2004}) alongside free-text corpora.

The main contributions of this work are as follows:
\begin{itemize}
    \item We propose a data-centric method for formulating the KG-based learning objectives of triple classification and entity/link prediction in the language modelling paradigm, and implement a framework for training transformers using the UMLS knowledge base in parallel with masked-language pre-training.
    \item Pre-training on the UMLS alongside the European Clinical Case Corpus \cite{minard_european_2021,magnini_e3c_2020}, we show that this method brings improvements to pre-trained models across a range of biomedical entity recognition tasks in three different languages, as well as functioning as a competitive pre-training strategy that requires much less training data in comparison to state-of-the-art transformer models.
    We release the monolingual and multilingual model weights trained in this way, UMLS-KGI\footnote{\textbf{K}nowledge \textbf{G}raph \textbf{I}ntegration}-BERT, as open-source resources for the clinical NLP research community.
    \item Based on this work, we release the Python library \texttt{bertify\_umls}, built mainly on the \texttt{transformers} and \texttt{pandas} libraries, which allows researchers to create custom text datasets and effectively use the UMLS knowledge base as a training corpus for BERT-style LMs.
\end{itemize}

\begin{figure*}
    \centering
    \includegraphics[scale=0.35]{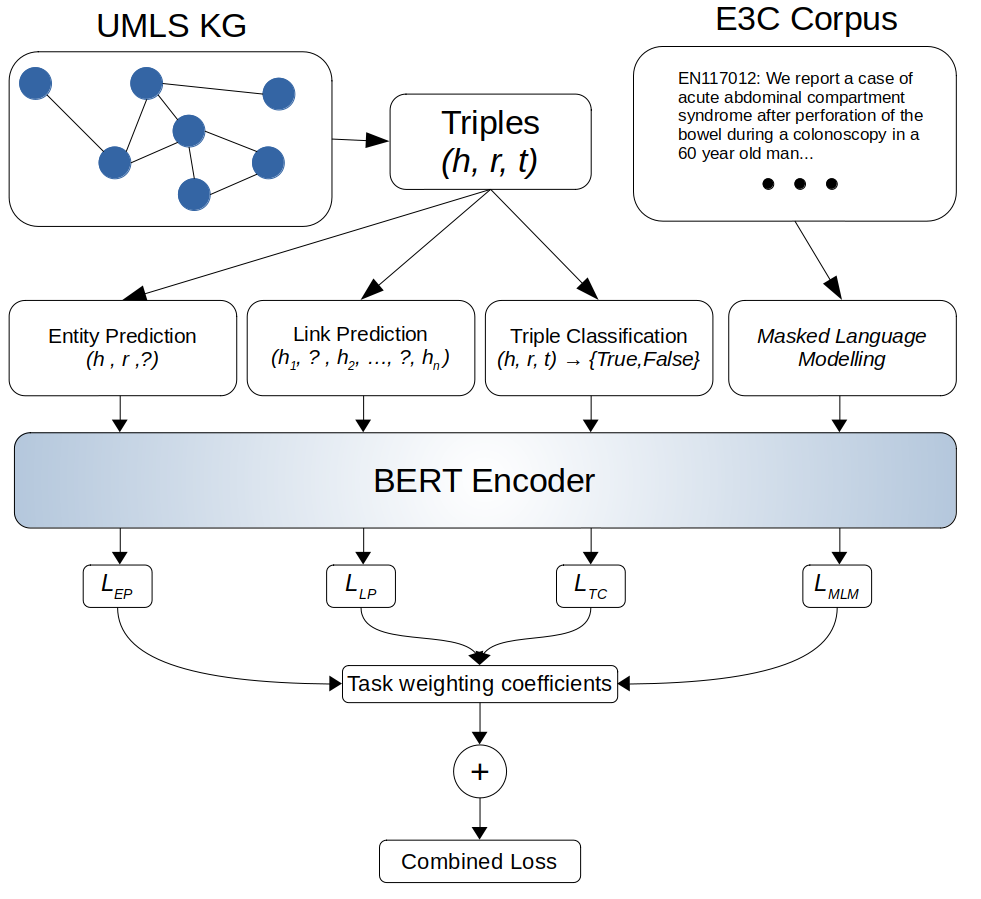}
    \caption{Overview of the UMLS-KGI pre-training process.}
    \label{fig:schema}
\end{figure*}

\section{Related Work}
\subsection{Pre-trained LMs for Medical Applications}
In general, the standard methodology for adapting neural text encoders to the biomedical domain has been to take a model that has been pre-trained on general-domain text corpora and continue this unsupervised pre-training on a medical corpus \cite{alrowili_biom-transformers_2021,lee_biobert_2020,alsentzer_publicly_2019}.
However, recent work has suggested that, given enough training data, it is preferable to pre-train these models on large domain-specific corpora only, without starting from a general-domain checkpoint \cite{gu_domain-specific_2021,rasmy_med-bert_2021}.
In this work we explore both approaches, extending existing biomedical and general-domain models as well as training BERT models from scratch on our own generated datasets.

\subsection{Knowledge-enhanced LMs}
Techniques for the incorporation of knowledge graph structure into BERT models can, broadly speaking, be divided into three categories, each focusing on one of the three fundamental components of a machine learning system, i.e. 1) the training data, 2) the model architecture and 3) the objective function to be optimised.
The first type of approach prioritises the augmentation of BERT's input data with information extracted from a knowledge graph.
This extra information can be numerical, e.g. pre-computed graph embeddings \cite{jeong_context-aware_2019} or textual, e.g. KG triples linked to input sentences \cite{liu_k-bert_2019}. 

The second type of approach focuses on adapting the architecture of BERT so that its language representations become fused with knowledge graph embeddings (KGEs) \cite{wang_kepler_2021,peters_knowledge_2019,zhang_ernie_2019}.
Knowledge graph fusion techniques such as these have been shown to be beneficial on certain English-language medical NLP tasks \cite{meng_mixture_2021,roy_incorporating_2021}.

Thirdly, the self-supervised pre-training objective of BERT models can be augmented using the kind of knowledge graph reasoning tasks used to build KGE models.
This approach is more commonly used for knowledge graph completion \cite{kim_multi-task_2020,yao_kgbert_2019} but has also been shown to be an effective strategy in the biomedical NLP domain \cite{hao_enhancing_2020}.

As previously mentioned, given that the medical domain is particularly exacting in terms of requirements for the use of structured facts, the exploration of ways in which ontological knowledge can be integrated into automated text processing is a very active area of research \cite{khosla_medfilter_2020,mondal_medical_2019}.
In particular, there have been multiple successful efforts to integrate the UMLS knowledge graph into BERT models, notably UmlsBERT \cite{michalopoulos_umlsbert_2021}, which proposes a data-augmentation technique allowing for concept and semantic type information to be linked to input text, and SapBERT \cite{liu_learning_2021,liu_self_2021}, which introduced a self-alignment strategy for learning from UMLS synonym pairs via a multi-similarity (MS) loss function to force related concepts closer to one another in BERT's representation space.
\citet{yuan_coder_2022} build on this strategy by applying MS loss to relation triples.
In contrast, in this work we show that information from the UMLS can be incorporated into BERT models in a simpler way, using only cross-entropy classification loss, while also balancing this training process with standard masked-language BERT pre-training.

Recent general overviews of the landscape of AI research have highlighted the importance of data-centric approaches to building models  \cite{zha_datacentric_2023,hamid_from_2022,jakubik_datacentric_2022} and in light of these trends this work focuses on types 1) and 3) of knowledge base integration described above, i.e. on improving the performance of standard model architectures by constructing high-quality datasets that can be integrated into the self-supervised language modelling paradigm by modifying the BERT objective function.
The motivation for this kind of approach is also to provide a pre-training framework that is more widely transferable and does not rely on any particular transformer-encoder architecture.

\subsection{Background: The UMLS Metathesaurus}

The UMLS Metathesaurus is a collection of biomedical taxonomies and vocabularies developed by the US National Library of Medicine.
It draws on a multitude of source vocabularies from thesauri to lists of controlled terms in a variety of healthcare-related domains.
The metathesaurus is organised by conceptual meaning, linking different terms and views for the same concept to each other and also identifying different kinds of relationships between concepts.

Terms and concepts are identified using three unique and permanent codes.
\begin{enumerate}
    \item The concept identifier (CUI), for each unique unit of meaning - different terms that refer to the same concept, e.g. \textit{Cardiac Arrythmia} and \textit{Irregular heartbeat}, will have the same CUI.
    \item The string identifier (SUI), for each unique term - strings that can have multiple meanings or interpretations, such as \textit{Cold}, can thus be associated with multiple different CUIs while retaining the same SUI.
    \item The atom identifier (AUI), for the basic building blocks of the knowledge graph: every occurrence of a string in any of the source vocabularies is assigned an AUI. 
\end{enumerate}

This work makes use of the 2022AB release, which contains 8,751,471 atoms defined by 3,711,072 CUIs, and 25,369,590 relations.
\begin{figure*}[]
\centering
	\includegraphics[scale=0.65]{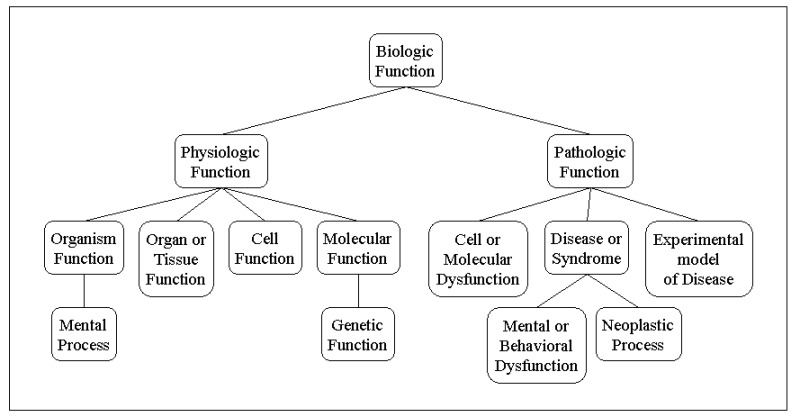}
	\caption{A portion of the UMLS semantic network - the \textit{biologic function} hierarchy. Figure taken from the UMLS Reference Manual.}
	\label{fig:semnet-example}
\end{figure*}

The data modelling approach implemented in this work also relies on the UMLS Semantic Network\footnote{\url{https://www.ncbi.nlm.nih.gov/books/NBK9679/}}, which provides a categorisation of the concepts in the metathesaurus and a set of relevant semantic relationships among these categories.
The network is made up of 127 semantic types, with groupings for organisms, anatomical structures, biologic function, chemicals, events, physical objects, and concepts or ideas (these are referred to as semantic groups); the aim being to provide a broad semantic interpretation framework for the range of different terminologies contained in the metathesaurus.
A portion of the hierarchical structure of the network, for the semantic type \textit{biologic function}, is illustrated in Figure \ref{fig:semnet-example} (see \url{https://www.ncbi.nlm.nih.gov/books/NBK9679/figure/ch05.F1/?report=objectonly}).

\section{Methodology}

\begin{table*}[]
    \centering
    \caption{Pre-training corpora sizes used in the experiments.}
    \footnotesize
    \begin{tabularx}{42.75em}{lcccccc}
        \label{tab:pretraindatasets}
        & \thead{Triple\\Classification} & \thead{Entity\\Prediction} & \thead{Paths} & \thead{E3C corpus\\(num. documents)} & \thead{Total Training\\Examples} & \thead{Memory\\Footprint} \\ \hline
        French & 200K & 100K & 64,208 & 25,740 & 389,948 & 604MB \\
        Spanish & 200K & 100K & 100K & 1,876 & 401,876 & 162MB \\
        English & 200K & 100K & 100K & 9,779 & 409,779 & 174MB \\ \hline
        Total   & 600K & 300K & 264,208 & 37,395 & 1,201,603 & 940MB \\ \hline
    \end{tabularx}
\end{table*}

In this work, we experiment with training BERT language models with three knowledge graph reasoning tasks derived from the UMLS, in addition to the standard masked-language modelling objective: entity prediction, link prediction and triple classification.
\subsection{Dataset Construction}
Formally, we consider the UMLS KG in the standard fashion, as a directed graph $G=(C,E,R)$ where $C$ is the set of all medical  concepts in the KG, $E$ the set of all edges or relations that link these concepts to one another, and $R$ the set of possible relation types, i.e. the labels $r$ for each $e\in E$.
The training sequences are thus generated from the KG dataset of ordered triples $(h,r,t)$ where $(h,r)\in C\times C$ and $r\in R$.
As a compendium of multiple different sources of taxonomic biomedical information, the UMLS metathesaurus contains multiple levels of granularity at which meaning representation can be analysed.
We consider three such levels of granularity in our work:
\begin{itemize}
    \item Terms - string descriptors for conceptual entities
    \item CUIs (Concept Unique Identifiers) - the basic unit of meaning representation for the nodes in the knowledge graph, i.e. the elements of the set $C$.
    \item Semantic Groups - these are groupings of concepts that can be considered to define the type of entity a concept represents; e.g. anatomical structure, chemical, disorder etc.
\end{itemize}
Each concept (CUI) can be associated with multiple terms and multiple semantic groups.
Thus, given that the entities $h$ and $t$ that make up the knowledge graph triples are represented as CUIs, in order to represent them as input text sequences for BERT models, we use the ``preferred term'' strings associated with the concepts $h$ and $t$, except in the case of synonym relations where we randomly select another of the terms associated with the concept in question to associate with $t$.
We also introduce a set of special tokens to represent the relation types $R$, of which there are seven (parent, child, synonymy, allowed qualifier, qualified by, broader, narrower).
Concretely, the tokenization function for BERT models forms text classification sequences from triples in the following way;
\begin{multline}
    \text{Tokenize}(h,r,t)=\texttt{[CLS]}w^h_1\cdots w^h_m \\ \texttt{[REL]}w^t_1\cdots w^t_n\texttt{[SEP]}
\end{multline}
where the $w_i$ represent the token sequences corresponding to the strings $h$ and $t$, \texttt{[CLS]} and \texttt{[SEP]} are BERT's standard classification and sequence-separation tokens as defined by \citet{devlin_bert_2019}, and \texttt{[REL]} is one of the relation tokens.
For link prediction, we construct a dataset of variable-length paths through the KG by iteratively selecting a list of triples $(h_1,r_1,t_1),\ldots,(h_n,r_n,t_n)$ where $h_{i+1}=t_i$ to form a path $p=(h_1,r_1,h_2,\ldots,r_n,t_n)$.  

\paragraph{Entity Prediction} The entity classification task can be trivially integrated into the masked-language objective of BERT, by masking the tokens associated with the concept $t$.
\paragraph{Link Prediction} We formulate link prediction as a narrow masked-language task by masking the relation tokens in the path dataset with another \textit{hidden relation} token, for which the model is trained to fill in one of six relation types - as the triple classification and entity prediction tasks already have the partial goal of improving the model's capability to associate synonymous terms with each other, we exclude synonym relations from the path dataset.
\paragraph{Triple Classification} Following the work of \citet{hao_enhancing_2020}, the triple classification objective is formulated as a binary classification problem where the model is tasked with classifying triples as true or false.
In order to generate training examples of false triples, we use two different negative sampling strategies.
Firstly, to provide directly contrastive examples for existing relations, we sample triples $(h,r,t)$ where $h$ and $t$ belong to different semantic groups and construct corresponding false triples with the same relation type and semantic group categories, i.e. $(\hat{h},r,\hat{t})\notin G$ where $\hat{h}$ and $\hat{t}$ are of the same semantic group as $h$ and $t$ respectively.
Secondly, to provide contrastive examples for relation types, we sample triples for which $h$ and $t$ are of the same semantic group, and form the negative training example by changing the relation type $r$.
To ensure balance, the triple classification datasets used in this work are made up of 50\% positive examples (real triples from the KG), 25\% examples generated by the first negative sampling method and the rest by the second.

We perform stratified sampling on the base knowledge graph according to semantic groups, i.e. we ensure that the proportional representation of each semantic group in the knowledge-base triples for each language is maintained in the training datasets. 
\paragraph{Mixed Objective Function}
In order to train BERT models using the UMLS-based reasoning tasks described above alongside the masked-language objective, each training example is augmented with an indicator label that tells the model which loss function to apply to the sequence in question.
The overall loss function is then calculated as
\begin{equation}
\label{loss}
    \mathcal{L}=\mathcal{L}_{\text{MLM}}+\alpha_1\mathcal{L}_{\text{EP}}+\alpha_2\mathcal{L}_{\text{LP}}+\alpha_3\mathcal{L}_{\text{TC}}
\end{equation}
where the $\alpha_i$ are scalar task-weighting coefficients and $\mathcal{L}_{\text{MLM}}$, $\mathcal{L}_{\text{EP}}$, $\mathcal{L}_{\text{LP}}$, and $\mathcal{L}_{\text{TC}}$ correspond to the loss values for masked language modelling, entity prediction, link prediction and triple classification respectively.
We use the standard cross-entropy classification loss for all tasks.

\section{Experiments}
For the evaluation of the approach described in the previous section, we restrict our attention in this paper to NER tasks.
Where possible, we use the datasets and training-evaluation-test splits that are publicly available via the Huggingface datasets library\footnote{\url{https://huggingface.co/datasets}}.
\subsection{KG-integrated pre-training}
\paragraph{Pre-training corpora} As a resource for masked-language pre-training, we utilise the European Clinical Case Corpus (E3C) version 2.0.0\footnote{\url{https://live.european-language-grid.eu/catalogue/corpus/7618}}, a freely-available multilingual corpus of clinical narratives.
We evaluate our method in three different languages; English, French and Spanish.
These languages were chosen as they are the three most well-represented languages in the metathesaurus for which we have access to pre-trained clinical BERT models for comparison.
The sizes of the combined UMLS-E3C datasets used are shown in Table \ref{tab:pretraindatasets}.

For each language, we compare the performance of 1) a transformer model trained from scratch on each monolingual dataset (KGI-BERT$_{EN,FR,ES}$) against 2) a multilingual version of the same model trained on all three datasets (KGI-BERT$_{m}$), 3) a pre-trained monolingual biomedical model and 4) the same pre-trained model with supplementary training on the corresponding monolingual UMLS-E3C dataset.

The UMLS-KGI models were trained for 64 epochs on each dataset, using the PyTorch implementation of the weighted ADAM optimizer \cite{loshchilov_decoupled_2019} with default parameters.
We use a maximal sequence length of 256 for the masked-language modelling sequences, an effective batch size of 1500 and a triangular learning rate schedule peaking at $7.5\times10^{-4}$.
To take into account the varying sizes of the components of the pre-training dataset we set the values of the coefficients of the loss function such that they are inversely proportional to the number of documents available: $$\alpha_i=\frac{\sum_{j=0,j\neq i}^3n_j}{2\sum_{k=0}^3n_k}$$ where the $n_k$ correspond to the number of documents in the training set for each UMLS-based task.
In this way, the E3C masked-language loss has the same weighting as the UMLS-based task losses.
\paragraph{Pre-trained models} For supplementary training, we make use of what are, to the best of our knowledge, the overall best-performing biomedical BERT models of their size (pre-trained using masked-language tasks only) for each language, according to baseline experiments on the NER tasks.

For French, we use DrBERT \cite{labrak_drbert_2023}, for Spanish the RoBERTa-based biomedical model released by \citet{carrino_biomedical_2021}, which we refer to as BioRoBERTa-ES, and for English PubMedBERT \cite{gu_domain-specific_2021}.
For training from scratch, we use the DistilBERT model configuration \cite{sanh_distilbert_2019} with 12 encoder layers and 12 attention heads.

\begin{table*}[]
    \centering
    \caption{Results on the French-language NER tasks. \textbf{Bold:} best result, {\ul underlined:} next best.}
    \footnotesize
    \begin{tabularx}{50em}{lcccccccccccc}
    \label{fr_results}
        & \multicolumn{3}{c}{\textbf{CAS-POS}} & \multicolumn{3}{c}{\textbf{CAS-SG}} & \multicolumn{3}{c}{\textbf{QUAERO-MEDLINE}} & \multicolumn{3}{c}{\textbf{ESSAI-POS}} \\ \cline{2-13}
        \textbf{Model} & \textbf{P} & \textbf{R} & \textbf{F1} & \textbf{P} & \textbf{R} & \textbf{F1} & \textbf{P} & \textbf{R} & \textbf{F1} & \textbf{P} & \textbf{R} & \textbf{F1} \\ \hline
        DrBERT-4GB & 90.94 & 91.59 & 90.84 & 65.86 & 64.89 & 62.20 & 68.65 & 69.38 & 66.66 & 94.83 & 95.08 & 94.69 \\
        + UMLS-KGI & {\ul 93.15} & {\ul 93.22} & {\ul 92.84} & 70.82 & \textbf{69.98} & {\ul 67.14} & 71.59 & 72.37 & 69.90 & 94.92 & 94.76 & 94.59 \\ \hline
        KGI-BERT$_{FR}$ & 88.55 & 88.40 & 87.82 & \textbf{71.57} & 66.90 & 65.79 & {\ul 71.78} & \textbf{72.93} & {\ul 70.75} & {\ul 95.46} & {\ul 95.40} & {\ul 95.18} \\
        KGI-BERT$_{m}$ & 90.87 & 90.58 & 90.16 & {\ul 71.14} & {\ul 69.81} & \textbf{67.28} & \textbf{72.04} & {\ul 72.89} & \textbf{70.96} & 94.88 & 94.84 & 94.55 \\ \hline
        SAPBERT-XL & \textbf{94.76} & \textbf{95.01} & \textbf{94.62} & 47.50 & 42.60 & 40.20 & 66.60 & 66.52 & 63.61 & \textbf{96.09} & \textbf{96.17} & \textbf{95.94} 
    \end{tabularx}
\end{table*}

\begin{table*}[]
    \centering
    \caption{Results on the English-language NER tasks.}
    \footnotesize
    \begin{tabularx}{39.5em}{lccccccccc}
    \label{en_results}
        & \multicolumn{3}{c}{\textbf{NCBI-Disease}} & \multicolumn{3} {c}{\textbf{BioRED-NER}} & \multicolumn{3}{c}{\textbf{JNLPBA04}} \\ \cline{2-10} \textbf{Model} & \textbf{P} & \textbf{R} & \textbf{F1} & \textbf{P} & \textbf{R} & \textbf{F1} & \textbf{P} & \textbf{R} & \textbf{F1} \\ \hline
        PubMedBERT & {\ul 93.81} & {\ul 94.26} & {\ul 93.53} & \textbf{84.76} & 85.33 & {\ul 83.35} & 81.57 & 82.59 & 81.13 \\
        + UMLS-KGI & \textbf{94.65} & \textbf{95.11} & \textbf{94.46} & {\ul 84.28} & \textbf{85.92} & \textbf{83.64} & \textbf{85.75} & \textbf{86.04} & \textbf{85.15}  \\ \hline
        KGI-BERT$_{EN}$ & 89.33 & 89.43 & 88.99 & 82.98 & {\ul 85.89} & 82.99 & {\ul 81.82} & {\ul 82.90} & {\ul 82.02} \\
        KGI-BERT$_{m}$ & 89.40 & 90.04 & 89.16 & 82.67 & 84.63 & 81.97 & 81.24 & 82.47 & 81.47 \\ \hline
        SAPBERT-XL & 93.65 & 94.15 & 93.40 & 65.01 & 66.65 & 62.05 & 80.44 & 81.31 & 79.91
    \end{tabularx}
\end{table*} 

\begin{table*}[]
    \centering
    \caption{Results on the Spanish-language NER tasks. \textbf{Bold:} best result, {\ul underlined:} next best.}
    \footnotesize
    \begin{tabularx}{29em}{lcccccc}
    \label{es_results}
        & \multicolumn{3}{c}{\textbf{PharmaCoNER}} & \multicolumn{3}{c}{\textbf{MEDDOCAN}} \\ \cline{2-7} 
        \textbf{Model} & \textbf{P} & \textbf{R} & \textbf{F1} & \textbf{P} & \textbf{R} & \textbf{F1} \\ \hline
        BioRoberta-ES  & 81.11  & 81.99  & 80.41  & 91.41  & 93.15 & 91.84 \\
        + UMLS-KGI & {\ul 83.52} & {\ul 84.30} & {\ul 83.90} & {\ul 93.65} & \textbf{95.32} & {\ul 91.99}  \\ \hline
        KGI-BERT$_{ES}$  & 79.95  & 80.14  & 78.11  & 92.28  & 92.93  & 92.17 \\
        KGI-BERT$_{m}$ & \textbf{85.05} & \textbf{85.95} & \textbf{85.49} & 92.32 & 92.65 & 91.98 \\ \hline
        SAPBERT-XL & 82.70 & 80.51 & 80.53 & \textbf{93.92} & {\ul 94.90} & \textbf{94.05}
    \end{tabularx}
\end{table*}

\subsection{Evaluation corpora}
We evaluate these models on nine different clinical entity recognition tasks; four in French, two in Spanish and three in English.
In order to ensure a fair comparison between models and evaluate more directly the knowledge transfer capabilities of the pre-trained models, we restrict ourselves to a \textit{one-shot} setting for all tasks, i.e. the model is given a single pass over the training data before being evaluated on the test set.
For all fine-tuning runs, we use an effective batch size of 4 (we found that very frequent optimizer updates give better results in for few-shot learning), learning rate $2\times10^{-5}$ and weight decay of 0.01.
\paragraph{CAS/ESSAIS} CAS \cite{grabar_cas_2018} and ESSAIS \cite{dalloux_supervised_2021} are corpora of clinical cases in French for which a subset is annotated with part-of-speech tags as well as semantic biomedical annotations (UMLS concepts, negation, and uncertainty).
We evaluate our models on the two corresponding medical POS-tagging tasks, CAS-POS and ESSAI-POS, as well as formulating a semantic-group token classification task using the CAS corpus annotations (CAS-SG).
\paragraph{QUAERO} The QUAERO French Medical Corpus \cite{neveol_quaero_2014} is a corpus of biomedical documents from EMEA and Medline annotated with UMLS concepts to facilitate entity recognition and document classification tasks.
The NER evaluation task we make use of here, QUAERO-MEDLINE, involves semantic group identification in the Medline documents.
\paragraph{PharmaCoNER} \cite{gonzalez_pharmaconer_2019} Designed for the automated recognition of pharmacological substances, compounds and proteins in Spanish-language clinical documents, this is a manually annotated subset of the Spanish Clinical Case Corpus (SPACCC \cite{intxaurrondo_spaccc_2018}).
\paragraph{MEDDOCAN} Similarly to PharmaCoNER, the MEDDOCAN corpus \cite{marimon_automatic_2019} is an annotated subset of SPACCC, in this case with semantic entity types relevant to clinical document anonymisation, i.e. words and expressions constituting Personal Health Information (PHI).
\paragraph{NCBI-Disease} \cite{dougan_ncbi_2014} The NCBI disease corpus is made up of PubMed abstracts with annotated disease mentions. In this work, we restrict our attention to token classification at the mention level. 
\paragraph{BioRED} \cite{luo_biored_2022} This corpus is designed for biomedical relation extraction and entity recognition; we focus on the latter in this work.
This task can be considered a more semantically general version of the NCBI disease recognition task, in that the BioRED corpus consists of PubMed abstracts annotated with a diverse range of entity types including genes, proteins and chemicals.
\paragraph{JNLPBA04 NER Dataset} \cite{collier_introduction_2004} Developed in the context of a biomedical entity recognition shared task, this corpus consists of Medline documents annotated with mentions of DNA, RNA, proteins, cell types and cell lines.

\subsubsection{Results}
We report the macro-averaged precision, recall and F1-score for each task.
These metrics are calculated on each sequence in the test set and then averaged across all training batches.
This evaluation setup is designed to allow for continuous monitoring of evaluation metrics \textit{during} training, and is in fact less than optimal for post-hoc testing, as the overall average can change based on the batch size chosen for evaluation.
This results in imbalanced precision and recall scores 

Results for the French, English and Spanish tasks can be seen in Tables \ref{fr_results}, \ref{en_results}, and \ref{es_results} respectively.
We find that the best-performing models are in general the pre-trained checkpoints for which training has been extended via knowledge graph integration.
This is unsurprising given that these are the models that have undergone the most domain-specific pre-training among all variants.
It is important to highlight, moreover, the fact that the KGI-BERT variants are competitive with the pre-trained baselines for many tasks, despite being trained on less data.
The largest improvements brought about by the UMLS-KGI training strategy can be seen in the French and Spanish tasks, suggesting that this technique will be more beneficial for lower-resource languages for which there is more room for improvement with respect to existing models.
\par We also include a comparison with the cross-lingual SAPBERT model \cite{liu_learning_2021}, which we include as a reference standard for UMLS-augmented BERT models.

The number of documents and target label classes for each evaluation task is show in Table \ref{eval_datasets}.
\begin{table}[]
    \centering
    \caption{Number of documents and target classes in the NER evaluation datasets}
    \resizebox{0.5 \textwidth}{!}{\begin{tabular}{lcccc}
    \label{eval_datasets}
        \textbf{Dataset} & \textbf{Train} & \textbf{Dev} & \textbf{Test} & \textbf{N. Classes} \\ \hline
        CAS-POS & 2,652 & 569 & 569 & 31 \\
        CAS-SG & 167 & 54 & 54 & 8 \\
        QUAERO-MEDLINE & 788 & 790 & 787 & 7 \\
        ESSAI-POS & 5,072 & 1,088 & 1,087 & 34 \\
        NCBI-Disease & 5,433 & 924 & 941 & 3 \\
        BioRED-NER & 387 & 98 & 97 & 7 \\
        JNLPBA04 & 16,619 & 1,927 & 3,856 & 11 \\
        PharmaCoNER & 500 & 250 & 250 & 5 \\
        MEDDOCAN & 500 & 250 & 250 & 22 \\ \hline
    \end{tabular}}
\end{table}

\subsection{Ablation Experiments}
In order to measure the relative effect of the three KG-derived pre-training tasks on downstream performance, we perform ablation experiments with the continually pre-trained models.
This involved comparing the downstream performance on the NER tasks of different versions of the UMLS-extended models, each with one of the three KG-based pre-training tasks excluded from the pre-training process.
For ablation, we use identical experimental settings to those described previously, except with 32 pre-training epochs rather than 64.

In general, the ablation results, for which the macro F1 scores are shown in Table \ref{tab:ablation}, suggest that the majority of the benefits in terms of NER performance are brought about by the link prediction task, although there are not enough statistically significant differences among the results to fully justify this conclusion.

It is clear also that certain tasks tend to add unhelpful noise to the model with respect to some tasks, in particular the ESSAI-POS task in French and the MEDDOCAN task in Spanish.
This may be due to the nature of these entity recognition tasks being more linked to general semantic patterns (i.e. parts-of-speech and identifying information) such that the addition of biomedical knowledge to the models does not improve their representation of the relevant concepts.

\begin{table*}[]
    \centering
    \caption{Macro-F1 scores for the ablation experiments.}
    \footnotesize
    \begin{tabularx}{45.75em}{llcccc}
    \label{tab:ablation}
        & & \multicolumn{4}{c}{\textbf{Dataset}} \\ \cline{3-6} 
        \textbf{Base Model} & \textbf{KG Tasks} & \textbf{CAS-POS} & \textbf{CAS-SG} & \textbf{QUAERO-MEDLINE} & \textbf{ESSAI-POS} \\ \hline
        DrBERT-4GB & - & 90.84 & 62.20 & 66.66 & \textbf{94.69} \\
        & EP+LP & 91.59 & 64.85 & 66.08 & 94.62 \\
        & EP+TC & 90.86 & 62.11 & 66.75 & {\ul 94.88} \\
        & TC+LP & {\ul 92.01} & {\ul 65.98} & {\ul 66.89} & 94.41 \\
        & all & \textbf{92.04} & \textbf{66.22} & \textbf{67.15} & 94.50 \\ \cline{3-5}
        & & \textbf{NCBI-Disease} & \textbf{BioRED-NER} & \textbf{JNLPBA04} & \\ \cline{3-5}
        PubMedBERT & - & 93.53 & 83.35 & 81.13 & \\
        & EP+LP & 93.24 & 82.40 & 81.25 & \\
        & EP+TC & 93.37 & 83.09 & 82.66 & \\
        & TC+LP & \textbf{94.13} & {\ul 83.38} & {\ul 84.30} & \\
        & all & {\ul 94.11} & \textbf{83.45} & \textbf{84.36} & \\ \cline{3-4}
        & & \textbf{PharmaCoNER} & \textbf{MEDDOCAN} & & \\ \cline{3-4}
        BioRoberta-ES & - & 81.11 & {\ul 91.84} & & \\
        & EP+LP & 81.12 & \textbf{91.86} & & \\
        & EP+TC & 82.40 & 91.80 & & \\
        & TC+LP & {\ul 83.22} & 91.71 & & \\
        & all & \textbf{83.46} & 91.77 & & \\ \hline
    \end{tabularx}
\end{table*}

\section{Conclusions and Future Work}
This paper introduces UMLS-KGI, a framework for training BERT models using knowledge graphs requiring highly minimal adjustments to the standard language modelling paradigm.
We show the potential of this method to increase the performance of BERT models on various NER tasks.
The results presented in this paper suggest that for clinical NER tasks, high-quality small-scale datasets derived from structured information, alongside alongside relatively small clinical text corpora, can be as effective as large-scale corpora for pre-training BERT models.
We make our models and data-processing pipelines freely available online.

Future work in this direction will involve the incorporation of more diverse graph-based reasoning tasks in the pre-training strategy with more fine-grained representation of relation types, as well as intrinsic evaluation of the UMLS-KGI-BERT language representations via embedding visualisation and interpretability studies.


\section*{Limitations}
The work presented in this paper is subject to a number of limitations which will be addressed in future work.
Firstly, we evaluate UMLS-KGI-BERT on a very narrow range of tasks limited to token classification - a broader range of information extraction and reasoning tasks would be necessary for a more complete picture of the utility of our pre-training methods.
In addition, we only train models for mid-to-high-resource languages; to properly validate the applicability of this approach, in particular the lessening of the need to rely on large training corpora, it will be necessary to train and evaluate such models in more low-resource settings.


\section*{Acknowledgements}
The experimentation in this work benefited from compute resources provided by GENCI-IDRIS (grant AD011013535R1).

\bibliography{anthology}

\begin{thebibliography}{51}
\expandafter\ifx\csname natexlab\endcsname\relax\def\natexlab#1{#1}\fi

\bibitem[{Alrowili and Shanker(2021)}]{alrowili_biom-transformers_2021}
Sultan Alrowili and Vijay Shanker. 2021.
\newblock \href {https://doi.org/10.18653/v1/2021.bionlp-1.24}
  {{BioM}-{Transformers}: {Building} {Large} {Biomedical} {Language} {Models}
  with {BERT}, {ALBERT} and {ELECTRA}}.
\newblock In \emph{Proceedings of the 20th {Workshop} on {Biomedical}
  {Language} {Processing}}, pages 221--227, Online. Association for
  Computational Linguistics.

\bibitem[{Alsentzer et~al.(2019)Alsentzer, Murphy, Boag, Weng, Jindi, Naumann,
  and McDermott}]{alsentzer_publicly_2019}
Emily Alsentzer, John Murphy, William Boag, Wei-Hung Weng, Di~Jindi, Tristan
  Naumann, and Matthew McDermott. 2019.
\newblock \href {https://doi.org/10.18653/v1/W19-1909} {Publicly {Available}
  {Clinical} {BERT} {Embeddings}}.
\newblock In \emph{Proceedings of the 2nd {Clinical} {Natural} {Language}
  {Processing} {Workshop}}, pages 72--78, Minneapolis, Minnesota, USA.
  Association for Computational Linguistics.

\bibitem[{Bodenreider(2004)}]{umls_2004}
O.~Bodenreider. 2004.
\newblock \href {https://doi.org/doi: 10.1093/nar/gkh061.} {The unified medical
  language system ({UMLS}): integrating biomedical terminology.}
\newblock PubMed PMID: 14681409; PubMed Central PMCID: PMC308795.

\bibitem[{Carrino et~al.(2021)Carrino, Armengol{-}Estap{\'{e}},
  Guti{\'{e}}rrez{-}Fandi{\~{n}}o, Llop{-}Palao, P{\`{a}}mies,
  Gonzalez{-}Agirre, and Villegas}]{carrino_biomedical_2021}
Casimiro~Pio Carrino, Jordi Armengol{-}Estap{\'{e}}, Asier
  Guti{\'{e}}rrez{-}Fandi{\~{n}}o, Joan Llop{-}Palao, Marc P{\`{a}}mies, Aitor
  Gonzalez{-}Agirre, and Marta Villegas. 2021.
\newblock \href {http://arxiv.org/abs/2109.03570} {Biomedical and clinical
  language models for spanish: On the benefits of domain-specific pretraining
  in a mid-resource scenario}.
\newblock \emph{CoRR}, abs/2109.03570.
\newblock ArXiv preprint 2109.03570.

\bibitem[{Chang et~al.(2020)Chang, Balažević, Allen, Chawla, Brandt, and
  Taylor}]{chang_benchmark_2020}
David Chang, Ivana Balažević, Carl Allen, Daniel Chawla, Cynthia Brandt, and
  Andrew Taylor. 2020.
\newblock \href {https://doi.org/10.18653/v1/2020.bionlp-1.18} {Benchmark and
  {Best} {Practices} for {Biomedical} {Knowledge} {Graph} {Embeddings}}.
\newblock In \emph{Proceedings of the 19th {SIGBioMed} {Workshop} on
  {Biomedical} {Language} {Processing}}, pages 167--176, Online. Association
  for Computational Linguistics.

\bibitem[{Collier and Kim(2004)}]{collier_introduction_2004}
Nigel Collier and Jin-Dong Kim. 2004.
\newblock \href {https://aclanthology.org/W04-1213} {Introduction to the
  bio-entity recognition task at {JNLPBA}}.
\newblock In \emph{Proceedings of the International Joint Workshop on Natural
  Language Processing in Biomedicine and its Applications
  ({NLPBA}/{B}io{NLP})}, pages 73--78, Geneva, Switzerland. COLING.

\bibitem[{Dalloux et~al.(2021)Dalloux, Claveau, Grabar, Oliveira, Moro, Gumiel,
  and Carvalho}]{dalloux_supervised_2021}
Clément Dalloux, Vincent Claveau, Natalia Grabar, Lucas Emanuel~Silva
  Oliveira, Claudia Maria~Cabral Moro, Yohan~Bonescki Gumiel, and
  Deborah~Ribeiro Carvalho. 2021.
\newblock \href {https://doi.org/10.1017/S1351324920000352} {Supervised
  learning for the detection of negation and of its scope in {F}rench and
  {B}razilian {P}ortuguese biomedical corpora}.
\newblock \emph{Natural Language Engineering}, 27(2):181–201.

\bibitem[{Devlin et~al.(2019)Devlin, Chang, Lee, and
  Toutanova}]{devlin_bert_2019}
Jacob Devlin, Ming-Wei Chang, Kenton Lee, and Kristina Toutanova. 2019.
\newblock \href {https://doi.org/10.18653/v1/N19-1423} {{BERT}: {Pre}-training
  of {Deep} {Bidirectional} {Transformers} for {Language} {Understanding}}.
\newblock In \emph{Proceedings of the 2019 {Conference} of the {North}
  {American} {Chapter} of the {Association} for {Computational} {Linguistics}:
  {Human} {Language} {Technologies}, {Volume} 1 ({Long} and {Short} {Papers})},
  pages 4171--4186, Minneapolis, Minnesota. Association for Computational
  Linguistics.

\bibitem[{Do{\u{g}}an et~al.(2014)Do{\u{g}}an, Leaman, and
  Lu}]{dougan_ncbi_2014}
Rezarta~Islamaj Do{\u{g}}an, Robert Leaman, and Zhiyong Lu. 2014.
\newblock {NCBI} disease corpus: a resource for disease name recognition and
  concept normalization.
\newblock \emph{Journal of biomedical informatics}, 47:1--10.

\bibitem[{Dubois et~al.(2017)Dubois, Romano, Jung, Shah, and
  Kale}]{dubois_effectiveness_2017}
S{\'e}bastien Dubois, Nathanael Romano, Kenneth Jung, Nigam~Haresh Shah, and
  David~C. Kale. 2017.
\newblock The effectiveness of transfer learning in electronic health records
  data.
\newblock In \emph{International Conference on Learning Representations}.

\bibitem[{Geiping and Goldstein(2022)}]{geiping_cramming_2022}
Jonas Geiping and Tom Goldstein. 2022.
\newblock \href {https://arxiv.org/abs/2212.14034} {Cramming: {Training} a
  {Language} {Model} on a {Single} {GPU} in {One} {Day}}.
\newblock \_eprint: 2212.14034.

\bibitem[{Gonzalez-Agirre et~al.(2019)Gonzalez-Agirre, Marimon, Intxaurrondo,
  Rabal, Villegas, and Krallinger}]{gonzalez_pharmaconer_2019}
Aitor Gonzalez-Agirre, Montserrat Marimon, Ander Intxaurrondo, Obdulia Rabal,
  Marta Villegas, and Martin Krallinger. 2019.
\newblock \href {https://doi.org/10.18653/v1/D19-5701} {Pharmaconer:
  Pharmacological substances, compounds and proteins named entity recognition
  track}.
\newblock In \emph{Proceedings of The 5th Workshop on BioNLP Open Shared
  Tasks}, pages 1--10, Hong Kong, China. Association for Computational
  Linguistics.

\bibitem[{Grabar et~al.(2018)Grabar, Claveau, and Dalloux}]{grabar_cas_2018}
Natalia Grabar, Vincent Claveau, and Clément Dalloux. 2018.
\newblock \href {https://doi.org/10.18653/v1/W18-5614} {{CAS}: {French}
  {Corpus} with {Clinical} {Cases}}.
\newblock In \emph{Proceedings of the {Ninth} {International} {Workshop} on
  {Health} {Text} {Mining} and {Information} {Analysis}}, pages 122--128,
  Brussels, Belgium. Association for Computational Linguistics.

\bibitem[{Gu et~al.(2021)Gu, Tinn, Cheng, Lucas, Usuyama, Liu, Naumann, Gao,
  and Poon}]{gu_domain-specific_2021}
Yu~Gu, Robert Tinn, Hao Cheng, Michael Lucas, Naoto Usuyama, Xiaodong Liu,
  Tristan Naumann, Jianfeng Gao, and Hoifung Poon. 2021.
\newblock \href {https://doi.org/10.1145/3458754} {Domain-{Specific} {Language}
  {Model} {Pretraining} for {Biomedical} {Natural} {Language} {Processing}}.
\newblock \emph{ACM Trans. Comput. Healthcare}, 3(1).
\newblock Place: New York, NY, USA Publisher: Association for Computing
  Machinery.

\bibitem[{Hamid(2022)}]{hamid_from_2022}
Oussama~H. Hamid. 2022.
\newblock \href {https://doi.org/10.1109/ITT56123.2022.9863935} {From
  model-centric to data-centric {AI}: A paradigm shift or rather a
  complementary approach?}
\newblock In \emph{2022 8th International Conference on Information Technology
  Trends ({ITT})}, pages 196--199.

\bibitem[{Hao et~al.(2020)Hao, Zhu, and Paschalidis}]{hao_enhancing_2020}
Boran Hao, Henghui Zhu, and Ioannis Paschalidis. 2020.
\newblock \href {https://doi.org/10.18653/v1/2020.coling-main.57} {Enhancing
  {Clinical} {BERT} {Embedding} using a {Biomedical} {Knowledge} {Base}}.
\newblock In \emph{Proceedings of the 28th {International} {Conference} on
  {Computational} {Linguistics}}, pages 657--661, Barcelona, Spain (Online).
  International Committee on Computational Linguistics.

\bibitem[{He et~al.(2022)He, Wang, Zhang, Li, Li, and Zeng}]{he_kgmttbert_2022}
Yong He, Cheng Wang, Shun Zhang, Nan Li, Zhaorong Li, and Zhenyu Zeng. 2022.
\newblock \href {http://arxiv.org/abs/2210.03970} {{KG-MTT-BERT}: Knowledge
  graph enhanced {BERT} for multi-type medical text classification}.
\newblock ArXiv preprint 2210.03970.

\bibitem[{Intxaurrondo(2018)}]{intxaurrondo_spaccc_2018}
Ander Intxaurrondo. 2018.
\newblock \href {https://doi.org/10.5281/zenodo.2560316} {{SPACCC}}.
\newblock {Funded by the Plan de Impulso de las Tecnologías del Lenguaje (Plan
  TL).}

\bibitem[{Jakubik et~al.(2022)Jakubik, Vössing, Kühl, Walk, and
  Satzger}]{jakubik_datacentric_2022}
Johannes Jakubik, Michael Vössing, Niklas Kühl, Jannis Walk, and Gerhard
  Satzger. 2022.
\newblock \href {https://doi.org/10.48550/arXiv.2212.11854} {Data-centric
  artificial intelligence}.
\newblock ArXiv preprint 2212.11854.

\bibitem[{Jeong et~al.(2019)Jeong, Jang, Shin, Park, and
  Choi}]{jeong_context-aware_2019}
Chanwoo Jeong, Sion Jang, Hyuna Shin, Eunjeong~Lucy Park, and Sungchul Choi.
  2019.
\newblock A context-aware citation recommendation model with bert and graph
  convolutional networks.
\newblock \emph{Scientometrics}, pages 1--16.

\bibitem[{Khosla et~al.(2020)Khosla, Vashishth, Lehman, and
  Rose}]{khosla_medfilter_2020}
Sopan Khosla, Shikhar Vashishth, Jill~Fain Lehman, and Carolyn Rose. 2020.
\newblock \href {https://doi.org/10.18653/v1/2020.emnlp-main.626}
  {{M}ed{F}ilter: {I}mproving {E}xtraction of {T}ask-relevant {U}tterances
  through {I}ntegration of {D}iscourse {S}tructure and {O}ntological
  {K}nowledge}.
\newblock In \emph{Proceedings of the 2020 Conference on Empirical Methods in
  Natural Language Processing (EMNLP)}, pages 7781--7797, Online. Association
  for Computational Linguistics.

\bibitem[{Kim et~al.(2020)Kim, Hong, Ko, and Seo}]{kim_multi-task_2020}
Bosung Kim, Taesuk Hong, Youngjoong Ko, and Jungyun Seo. 2020.
\newblock \href {https://doi.org/10.18653/v1/2020.coling-main.153}
  {Multi-{Task} {Learning} for {Knowledge} {Graph} {Completion} with
  {Pre}-trained {Language} {Models}}.
\newblock In \emph{Proceedings of the 28th {International} {Conference} on
  {Computational} {Linguistics}}, pages 1737--1743, Barcelona, Spain (Online).
  International Committee on Computational Linguistics.

\bibitem[{Labrak et~al.(2023)Labrak, Bazoge, Dufour, Rouvier, Morin, Daille,
  and Gourraud}]{labrak_drbert_2023}
Yanis Labrak, Adrien Bazoge, Richard Dufour, Mickael Rouvier, Emmanuel Morin,
  Béatrice Daille, and Pierre-Antoine Gourraud. 2023.
\newblock \href {http://arxiv.org/abs/2304.00958} {{DrBERT}: A robust
  pre-trained model in {F}rench for biomedical and clinical domains}.
\newblock ArXiv preprint 2304.00958.

\bibitem[{Lee et~al.(2020)Lee, Yoon, Kim, Kim, Kim, So, and
  Kang}]{lee_biobert_2020}
J.~Lee, W.~Yoon, S.~Kim, D.~Kim, S.~Kim, C.~Ho So, and J.~Kang. 2020.
\newblock {BioBERT}: a pre-trained biomedical language representation model for
  biomedical text mining.
\newblock \emph{Bioinformatics 2020}.

\bibitem[{Li et~al.(2021)Li, Pan, Goldwasser, Verma, Wong, Nuzumlalı, Rosand,
  Li, Zhang, Chang, Taylor, Krumholz, and Radev}]{li_neural_2021}
Irene Li, Jessica Pan, Jeremy Goldwasser, Neha Verma, Wai~Pan Wong,
  Muhammed~Yavuz Nuzumlalı, Benjamin Rosand, Yixin Li, Matthew Zhang, David
  Chang, R.~Andrew Taylor, Harlan~M. Krumholz, and Dragomir Radev. 2021.
\newblock \href {http://arxiv.org/abs/2107.02975} {Neural natural language
  processing for unstructured data in electronic health records: a review}.
\newblock ArXiv preprint 2107.02975.

\bibitem[{Li et~al.(2020)Li, Rao, Solares, Hassaine, Ramakrishnan, Canoy, Zhu,
  Rahimi, and Salimi-Khorshidi}]{li_behrt_2020}
Yikuan Li, Shishir Rao, José Roberto~Ayala Solares, Abdelaali Hassaine, Rema
  Ramakrishnan, Dexter Canoy, Yajie Zhu, Kazem Rahimi, and Gholamreza
  Salimi-Khorshidi. 2020.
\newblock \href {https://doi.org/10.1038/s41598-020-62922-y} {{BEHRT}:
  {Transformer} for {Electronic} {Health} {Records}}.
\newblock \emph{Scientific Reports}, 10(1):7155.
\newblock Number: 1 Publisher: Nature Publishing Group.

\bibitem[{Liu et~al.(2021{\natexlab{a}})Liu, Shareghi, Meng, Basaldella, and
  Collier}]{liu_self_2021}
Fangyu Liu, Ehsan Shareghi, Zaiqiao Meng, Marco Basaldella, and Nigel Collier.
  2021{\natexlab{a}}.
\newblock Self-alignment pretraining for biomedical entity representations.
\newblock In \emph{Proceedings of the 2021 Conference of the North American
  Chapter of the Association for Computational Linguistics: Human Language
  Technologies}, pages 4228--4238.

\bibitem[{Liu et~al.(2021{\natexlab{b}})Liu, Vuli{\'c}, Korhonen, and
  Collier}]{liu_learning_2021}
Fangyu Liu, Ivan Vuli{\'c}, Anna Korhonen, and Nigel Collier.
  2021{\natexlab{b}}.
\newblock Learning domain-specialised representations for cross-lingual
  biomedical entity linking.
\newblock In \emph{Proceedings of ACL-IJCNLP 2021}, pages 565--574.

\bibitem[{Liu et~al.(2019)Liu, Zhou, Zhao, Wang, Ju, Deng, and
  Wang}]{liu_k-bert_2019}
Weijie Liu, Peng Zhou, Zhe Zhao, Zhiruo Wang, Qi~Ju, Haotang Deng, and Ping
  Wang. 2019.
\newblock K-{BERT}: {Enabling} {Language} {Representation} with {Knowledge}
  {Graph}.
\newblock In \emph{{AAAI} {Conference} on {Artificial} {Intelligence}}.

\bibitem[{Loshchilov and Hutter(2019)}]{loshchilov_decoupled_2019}
Ilya Loshchilov and Frank Hutter. 2019.
\newblock \href {http://arxiv.org/abs/1711.05101} {Decoupled weight decay
  regularization}.
\newblock ArXiv:1711.05101.

\bibitem[{Luo et~al.(2022)Luo, Lai, Wei, Arighi, and Lu}]{luo_biored_2022}
Ling Luo, Po{-}Ting Lai, Chih{-}Hsuan Wei, Cecilia~N. Arighi, and Zhiyong Lu.
  2022.
\newblock \href {https://doi.org/10.48550/arXiv.2204.04263} {Biored: {A}
  comprehensive biomedical relation extraction dataset}.
\newblock \emph{CoRR}, abs/2204.04263.

\bibitem[{Magnini et~al.(2020)Magnini, Altuna, Lavelli, Speranza, and
  Zanoli}]{magnini_e3c_2020}
B.~Magnini, B.~Altuna, A.~Lavelli, M.~Speranza, and R.~Zanoli. 2020.
\newblock The e3c project: Collection and annotation of a multilingual corpus
  of clinical cases.

\bibitem[{Marimon et~al.(2019)Marimon, Gonzalez-Agirre, Intxaurrondo,
  Rodriguez, Martin, Villegas, and Krallinger}]{marimon_automatic_2019}
Montserrat Marimon, Aitor Gonzalez-Agirre, Ander Intxaurrondo, Heidy Rodriguez,
  Jose~Lopez Martin, Marta Villegas, and Martin Krallinger. 2019.
\newblock Automatic de-identification of medical texts in spanish: the meddocan
  track, corpus, guidelines, methods and evaluation of results.
\newblock In \emph{IberLEF@ SEPLN}, pages 618--638.

\bibitem[{Meng et~al.(2021)Meng, Liu, Clark, Shareghi, and
  Collier}]{meng_mixture_2021}
Zaiqiao Meng, Fangyu Liu, Thomas Clark, Ehsan Shareghi, and Nigel Collier.
  2021.
\newblock \href {https://doi.org/10.18653/v1/2021.emnlp-main.383}
  {Mixture-of-{Partitions}: {Infusing} {Large} {Biomedical} {Knowledge}
  {Graphs} into {BERT}}.
\newblock In \emph{Proceedings of the 2021 {Conference} on {Empirical}
  {Methods} in {Natural} {Language} {Processing}}, pages 4672--4681, Online and
  Punta Cana, Dominican Republic. Association for Computational Linguistics.

\bibitem[{Michalopoulos et~al.(2021)Michalopoulos, Wang, Kaka, Chen, and
  Wong}]{michalopoulos_umlsbert_2021}
George Michalopoulos, Yuanxin Wang, Hussam Kaka, Helen Chen, and Alexander
  Wong. 2021.
\newblock \href {https://doi.org/10.18653/v1/2021.naacl-main.139}
  {{U}mls{BERT}: Clinical domain knowledge augmentation of contextual
  embeddings using the {U}nified {M}edical {L}anguage {S}ystem
  {M}etathesaurus}.
\newblock In \emph{Proceedings of the 2021 Conference of the North American
  Chapter of the Association for Computational Linguistics: Human Language
  Technologies}, pages 1744--1753, Online. Association for Computational
  Linguistics.

\bibitem[{Minard et~al.(2021)Minard, Zanoli, Altuna, Speranza, Magnini, and
  Lavelli}]{minard_european_2021}
Anne-Lyse Minard, Roberto Zanoli, Begoña Altuna, Manuela Speranza, Bernardo
  Magnini, and Alberto Lavelli. 2021.
\newblock \href {https://doi.org/10.57771/DEY2-G751} {European clinical case
  corpus}.
\newblock Bruno Kessler Foundation.

\bibitem[{Mondal et~al.(2019)Mondal, Purkayastha, Sarkar, Goyal, Pillai,
  Bhattacharyya, and Gattu}]{mondal_medical_2019}
Ishani Mondal, Sukannya Purkayastha, Sudeshna Sarkar, Pawan Goyal, Jitesh
  Pillai, Amitava Bhattacharyya, and Mahanandeeshwar Gattu. 2019.
\newblock \href {https://doi.org/10.18653/v1/W19-1912} {Medical entity linking
  using triplet network}.
\newblock In \emph{Proceedings of the 2nd Clinical Natural Language Processing
  Workshop}, pages 95--100, Minneapolis, Minnesota, USA. Association for
  Computational Linguistics.

\bibitem[{Naseem et~al.(2022)Naseem, Bandi, Raza, Rashid, and
  Chakravarthi}]{naseem_incorporating_2022}
Usman Naseem, Ajay Bandi, Shaina Raza, Junaid Rashid, and Bharathi~Raja
  Chakravarthi. 2022.
\newblock \href {https://doi.org/10.18653/v1/2022.bionlp-1.10} {Incorporating
  {Medical} {Knowledge} to {Transformer}-based {Language} {Models} for
  {Medical} {Dialogue} {Generation}}.
\newblock In \emph{Proceedings of the 21st {Workshop} on {Biomedical}
  {Language} {Processing}}, pages 110--115, Dublin, Ireland. Association for
  Computational Linguistics.

\bibitem[{Névéol et~al.(2014)Névéol, Grouin, Leixa, Rosset, and
  Zweigenbaum}]{neveol_quaero_2014}
Aurélie Névéol, Cyril Grouin, Jeremy Leixa, Sophie Rosset, and Pierre
  Zweigenbaum. 2014.
\newblock The {QUAERO} {French} {Medical} {Corpus}: {A} {Ressource} for
  {Medical} {Entity} {Recognition} and {Normalization}.
\newblock In \emph{Proc of {BioTextMining} {Work}}, pages 24--30.

\bibitem[{Peters et~al.(2019)Peters, Neumann, Logan, Schwartz, Joshi, Singh,
  and Smith}]{peters_knowledge_2019}
Matthew~E. Peters, Mark Neumann, Robert Logan, Roy Schwartz, Vidur Joshi,
  Sameer Singh, and Noah~A. Smith. 2019.
\newblock \href {https://doi.org/10.18653/v1/D19-1005} {Knowledge enhanced
  contextual word representations}.
\newblock In \emph{Proceedings of the 2019 Conference on Empirical Methods in
  Natural Language Processing and the 9th International Joint Conference on
  Natural Language Processing (EMNLP-IJCNLP)}, pages 43--54, Hong Kong, China.
  Association for Computational Linguistics.

\bibitem[{Rasmy et~al.(2021)Rasmy, Xiang, Xie, Tao, and
  Zhi}]{rasmy_med-bert_2021}
Laila Rasmy, Yang Xiang, Ziqian Xie, Cui Tao, and Degui Zhi. 2021.
\newblock \href {https://doi.org/10.1038/s41746-021-00455-y} {Med-{BERT}:
  pretrained contextualized embeddings on large-scale structured electronic
  health records for disease prediction}.
\newblock \emph{npj Digital Medicine}, 4(1):86.

\bibitem[{Roy and Pan(2021)}]{roy_incorporating_2021}
Arpita Roy and Shimei Pan. 2021.
\newblock \href {https://doi.org/10.18653/v1/2021.emnlp-main.435}
  {Incorporating medical knowledge in {BERT} for clinical relation extraction}.
\newblock In \emph{Proceedings of the 2021 {Conference} on {Empirical}
  {Methods} in {Natural} {Language} {Processing}}, pages 5357--5366, Online and
  Punta Cana, Dominican Republic. Association for Computational Linguistics.

\bibitem[{Sanh et~al.(2019)Sanh, Debut, Chaumond, and
  Wolf}]{sanh_distilbert_2019}
Victor Sanh, Lysandre Debut, Julien Chaumond, and Thomas Wolf. 2019.
\newblock Distil{BERT}, a distilled version of bert: smaller, faster, cheaper
  and lighter.
\newblock \emph{ArXiv}, abs/1910.01108.

\bibitem[{Singhal et~al.(2022)Singhal, Azizi, Tu, Mahdavi, Wei, Chung, Scales,
  Tanwani, Cole-Lewis, Pfohl, Payne, Seneviratne, Gamble, Kelly, Scharli,
  Chowdhery, Mansfield, Arcas, Webster, Corrado, Matias, Chou, Gottweis,
  Tomasev, Liu, Rajkomar, Barral, Semturs, Karthikesalingam, and
  Natarajan}]{singhal_large_2022}
Karan Singhal, Shekoofeh Azizi, Tao Tu, S.~Sara Mahdavi, Jason Wei, Hyung~Won
  Chung, Nathan Scales, Ajay Tanwani, Heather Cole-Lewis, Stephen Pfohl, Perry
  Payne, Martin Seneviratne, Paul Gamble, Chris Kelly, Nathaneal Scharli,
  Aakanksha Chowdhery, Philip Mansfield, Blaise Aguera~y Arcas, Dale Webster,
  Greg~S. Corrado, Yossi Matias, Katherine Chou, Juraj Gottweis, Nenad Tomasev,
  Yun Liu, Alvin Rajkomar, Joelle Barral, Christopher Semturs, Alan
  Karthikesalingam, and Vivek Natarajan. 2022.
\newblock \href {http://arxiv.org/abs/2212.13138} {Large {Language} {Models}
  {Encode} {Clinical} {Knowledge}}.
\newblock ArXiv:2212.13138 [cs].

\bibitem[{Tiwari et~al.(2020)Tiwari, Colborn, Smith, Xing, Ghosh, and
  Rosenberg}]{tiwari_assessment_2020}
Premanand Tiwari, Kathryn~L. Colborn, Derek~E. Smith, Fuyong Xing, Debashis
  Ghosh, and Michael~A. Rosenberg. 2020.
\newblock \href {https://doi.org/10.1001/jamanetworkopen.2019.19396}
  {Assessment of a {Machine} {Learning} {Model} {Applied} to {Harmonized}
  {Electronic} {Health} {Record} {Data} for the {Prediction} of {Incident}
  {Atrial} {Fibrillation}}.
\newblock \emph{JAMA Network Open}, 3(1).
\newblock Publisher: American Medical Association.

\bibitem[{Vaswani et~al.(2017)Vaswani, Shazeer, Parmar, Uszkoreit, Jones,
  Gomez, Kaiser, and Polosukhin}]{vaswani_attention_2017}
Ashish Vaswani, Noam Shazeer, Niki Parmar, Jakob Uszkoreit, Llion Jones,
  Aidan~N Gomez, Lukasz Kaiser, and Illia Polosukhin. 2017.
\newblock \href
  {https://proceedings.neurips.cc/paper/2017/file/3f5ee243547dee91fbd053c1c4a845aa-Paper.pdf}
  {Attention is {All} you {Need}}.
\newblock In \emph{Advances in {Neural} {Information} {Processing} {Systems}},
  volume~30. Curran Associates, Inc.

\bibitem[{Wang et~al.(2021)Wang, Gao, Zhu, Zhang, Liu, Li, and
  Tang}]{wang_kepler_2021}
Xiaozhi Wang, Tianyu Gao, Zhaocheng Zhu, Zhengyan Zhang, Zhiyuan Liu, Juanzi
  Li, and Jian Tang. 2021.
\newblock \href {https://doi.org/10.1162/tacl_a_00360} {{KEPLER}: {A} {Unified}
  {Model} for {Knowledge} {Embedding} and {Pre}-trained {Language}
  {Representation}}.
\newblock \emph{Transactions of the Association for Computational Linguistics},
  9:176--194.
\newblock Place: Cambridge, MA Publisher: MIT Press.

\bibitem[{Yao et~al.(2019)Yao, Mao, and Luo}]{yao_kgbert_2019}
Liang Yao, Chengsheng Mao, and Yuan Luo. 2019.
\newblock \href {http://arxiv.org/abs/1909.03193} {{KG-BERT: BERT} for
  knowledge graph completion}.
\newblock ArXiv preprint 1909.03193.

\bibitem[{Yuan et~al.(2022)Yuan, Zhao, Sun, Li, Wang, and Yu}]{yuan_coder_2022}
Zheng Yuan, Zhengyun Zhao, Haixia Sun, Jiao Li, Fei Wang, and Sheng Yu. 2022.
\newblock \href {https://doi.org/10.1016/j.jbi.2021.103983} {{CODER}:
  {Knowledge}-infused cross-lingual medical term embedding for term
  normalization.}
\newblock \emph{Journal of biomedical informatics}, 126:103983.
\newblock Place: United States.

\bibitem[{Zha et~al.(2023)Zha, Bhat, Lai, Yang, Jiang, Zhong, and
  Hu}]{zha_datacentric_2023}
Daochen Zha, Zaid~Pervaiz Bhat, Kwei-Herng Lai, Fan Yang, Zhimeng Jiang,
  Shaochen Zhong, and Xia Hu. 2023.
\newblock \href {http://arxiv.org/abs/2303.10158} {Data-centric artificial
  intelligence: A survey}.

\bibitem[{Zhang et~al.(2019)Zhang, Han, Liu, Jiang, Sun, and
  Liu}]{zhang_ernie_2019}
Zhengyan Zhang, Xu~Han, Zhiyuan Liu, Xin Jiang, Maosong Sun, and Qun Liu. 2019.
\newblock \href {https://doi.org/10.18653/v1/P19-1139} {{ERNIE}: {Enhanced}
  {Language} {Representation} with {Informative} {Entities}}.
\newblock In \emph{Proceedings of the 57th {Annual} {Meeting} of the
  {Association} for {Computational} {Linguistics}}, pages 1441--1451, Florence,
  Italy. Association for Computational Linguistics.

\end{thebibliography}
\bibliographystyle{acl_natbib}

\appendix

\section{Dataset Statistics}
\label{sec:dataset_stats}
\subsection{UMLS Knowledge Graph}
We use the 2022AB release of the UMLS knowledge graph, which contains 8,751,471 concepts defined by 3,711,072 unique identifiers (CUIs), and 25,369,590 relations.
Restricting our attention to semantic types related to human biology and medicine, we end up with the base dataset outlined in Table \ref{umls_dataset}.

\begin{table}[]
    \centering
    \caption{Size of the UMLS dataset from which the KG-based pre-training corpus was sampled.}
    \begin{tabularx}{0.5 \textwidth}{lccc}
    \label{umls_dataset}
        \textbf{Language} & \textbf{Terms} & \textbf{CUIs} & \textbf{Relations} \\ \hline
        English & 3,912,195 & 2,245,468 & 17,121,829 \\
        Spanish & 303,978 & 118,061 & 437,578 \\
        French & 202,963 & 171,060 & 669,006 \\ \hline
        Total & 4,419,136 & 2,534,589 & 18,228,413
    \end{tabularx}
\end{table}

\section{Supplementary Experimental Details}
\paragraph{Pre-trained Checkpoints}
We use the following pre-trained model weights downloaded from the HuggingFace model hub as baseline models;
\begin{itemize}
    \item DrBERT: \texttt{Dr-BERT/DrBERT-4GB}
    \item PubMedBERT: \texttt{microsoft/BiomedNLP-PubMedBERT-base-\\uncased-abstract}
    \item BioRoBERTa-ES: \texttt{PlanTL-GOB-ES/roberta-base-\\biomedical-clinical-es}
\end{itemize}
\paragraph{Model Hyperparameters} The hyperparameter settings used for the pre-training on the UMLS-based dataset are shown in Table \ref{pt_params}.
The pre-training process used a linear learning rate schedule with warmup, where the learning rate increases from zero over the warmup period until it reaches the specified before decaying linearly over the rest of the training steps.
In the interest of minimising the energy consumption of our experiments, we carried out very minimal hyperparameter search, leaving most parameters at their default values.
The experiments were run using Python 3.8.15, with PyTorch version 2.0.0 and CUDA 11.8, along with the \texttt{transformers} library version 4.27.4.
\begin{table}[]
    \centering
    \caption{Hyperparameter settings for pre-training the UMLS-KGI models.}
    \begin{tabular}{lc}
    \label{pt_params}
        \textbf{Parameter}          & \textbf{Value} \\ \hline
        Sequence Length             & 256            \\
        Learning rate               & 0.00075        \\
        Learning rate warmup steps  & 10,770         \\
        Batch size                  & 15             \\
        Gradient accumulation steps & 100            \\
        MLM probability             & 0.15           \\ \hline
    \end{tabular}
\end{table}

\paragraph{Hardware specifications} The pre-training experiments were run on four Nvidia Tesla V100 GPUs with 32GB of RAM, while the fine-tuning experiments were run on an RTX 2080 Ti with 11GB of RAM.


\section{Additional Work: UMLS-KGI-v2}

The training process described in this paper took heavy inspiration from the ``cramming'' (single-GPU, limited-budget training) guidelines given by \cite{geiping_cramming_2022}.

As a continuation of these experiments, we moved on to the \texttt{v2} UMLS-KGI models, which were trained with a less constrained experimental setup.

For these models, which utilised 4-GPU training runs, an effective batch size of 1536 was used, and the training was extended to 200 epochs for the from-scratch models and 128 for the ones employing continual pre-training.
The \texttt{v2} training used a learning rate of $5\times10^{-5}$ with a linear warm-up over the first 30k steps.

In addition, while the \texttt{v1} models used an encoder architecture with 12 layers and 12 attention heads, the \texttt{v2} models have just 8 of each.
This decision was made due to observations that, for the scale of datasets used in this work, the $12\times12$ architecture is often under-trained and displays erratic downstream behaviour, while $8\times8$ models show better performance and higher levels of robustness.

Comparative results for these models are shown in Tables \ref{tab:fr-kgi-results}, \ref{tab:en-kgi-results}, and \ref{tab:es-kgi-results}.
\begin{table*}[ht]
    \centering
    \begin{tabular}{l|c|c|c|c}
        \textbf{Model} & \textbf{CAS-SG} & \textbf{QUAERO-MEDLINE} & \textbf{CAS-POS} & \textbf{ESSAI-POS}  \\ \hline
        \texttt{SapBERT-XL} & 40.20 & 63.61 & \textbf{94.62} & \textbf{95.94} \\ \hline
        \texttt{DrBERT-4GB} & 62.20 & 66.66 & 90.84 & \textit{94.69} \\
        + \texttt{UMLS-KGI-v1} & 67.14 & 69.90 & \textit{92.84} & 94.59 \\ 
        + \texttt{UMLS-KGI-v2} & \textit{68.94} & \textit{71.56} & 91.07 & 92.88 \\ \hline
        \texttt{KGI-BERT\textsubscript{FR}-v1} & 65.79 & 70.75 & 87.82 & \textit{95.18} \\
        \texttt{KGI-BERT\textsubscript{m}-v1} & \textit{67.28} & \textit{70.96} & 90.16 & 94.55 \\ \hline
        \texttt{KGI-BERT\textsubscript{FR}-v2} & 67.00 & 65.15 & 89.56 & 91.22\\
        \texttt{KGI-BERT\textsubscript{m}-v2} & \textbf{74.82} & \textbf{72.14} & \textit{91.35} & 93.59
    \end{tabular}
    \caption{Macro $F_1$ scores on the French-language token classification tasks.}
    \label{tab:fr-kgi-results}
\end{table*}

\begin{table*}[ht]
    \centering
    \begin{tabular}{l|c|c|c}
        \textbf{Model} & \textbf{NCBI-Disease} & \textbf{BioRED-NER} & \textbf{JNLPBA04} \\ \hline
        \texttt{SapBERT-XL} & 93.40 & 62.05 & 79.91 \\ \hline
        \texttt{PubMedBERT} & 93.53 & 83.35 & 81.13 \\
        + \texttt{UMLS-KGI-v1} & \textbf{94.46} & 83.64 & \textbf{85.15}  \\
        + \texttt{UMLS-KGI-v2} & \textit{94.43} & \textit{86.96} & 76.13 \\ \hline
        \texttt{KGI-BERT\textsubscript{EN}-v1} & 88.99 & 82.99 & \textit{82.02} \\
        \texttt{KGI-BERT\textsubscript{m}-v1} & 89.16 & 81.97 & \textit{81.47} \\ \hline
        \texttt{KGI-BERT\textsubscript{EN}-v2} & \textit{94.22} & \textbf{88.01} & 76.99 \\
        \texttt{KGI-BERT\textsubscript{m}-v2} & 93.99 & \textit{87.33} & 79.53
    \end{tabular}
    \caption{Macro $F_1$ scores on the English-language token classification tasks.}
    \label{tab:en-kgi-results}
\end{table*}

\begin{table*}[ht]
    \centering
    \begin{tabular}{l|c|c}
        \textbf{Model} & \textbf{PharmaCoNER} & \textbf{MEDDOCAN} \\ \hline
        \texttt{SapBERT-XL} & 80.53 & \textbf{94.05} \\ \hline
        \texttt{BioRoBERTa-ES}  & 80.41  & 91.84 \\
        + \texttt{UMLS-KGI-v1} & 83.90 & \textit{91.99}  \\
        + \texttt{UMLS-KGI-v2} & \textit{85.40} & 90.73 \\ \hline
        \texttt{KGI-BERT\textsubscript{ES}-v1} & 78.11 & \textit{92.17} \\
        \texttt{KGI-BERT\textsubscript{m}-v1} & \textit{85.49} & 91.98 \\ \hline
        \texttt{KGI-BERT\textsubscript{ES}-v2} & 83.88 & 85.00 \\
        \texttt{KGI-BERT\textsubscript{m}-v2} & \textbf{86.81} & 88.98
    \end{tabular}
    \caption{Macro $F_1$ scores on the Spanish-language token classification tasks.}
    \label{tab:es-kgi-results}
\end{table*}

\end{document}